\documentclass[journal]{ieeetran}

\usepackage{amsmath,amssymb,bm} 
\usepackage{epsfig}             
\usepackage{textcomp}           
\usepackage{multirow}           
\usepackage{array}              
\usepackage{xcolor}             
\usepackage{tikz}               
\usepackage{mathtools}          
\usepackage{color,soul}         
\usetikzlibrary{shapes,arrows}  

\makeatletter
\renewcommand*\env@matrix[1][*\c@MaxMatrixCols c]{%
  \hskip -\arraycolsep
  \let\@ifnextchar\new@ifnextchar
  \array{#1}}
\makeatother

\begin{document}

\tikzstyle{block} = [draw, fill=white, rectangle, 
    minimum height=2em, minimum width=2em]
\tikzstyle{sum} = [draw, fill=white, circle, node distance=1cm]
\tikzstyle{input} = [coordinate]
\tikzstyle{output} = [coordinate]
\tikzstyle{pinstyle} = [pin edge={to-,thin,black}]

\title{\LARGE\bf High-Precision Surgical Robotic System for Intraocular Procedures}
\author{Yu-Ting Lai$^1$, Jacob Rosen$^1$, Yasamin Foroutani$^1$, Ji Ma$^1$, Wen-Cheng Wu$^1$, \\Jean-Pierre Hubschman$^2$, and Tsu-Chin Tsao$^1$%
\thanks{This work was supported by U.S. NIH/R01EY030595 and  NIH/R01EY029689.}
\thanks{$^1$Yu-Ting Lai, Jacob Rosen, Yasamin Foroutani, Ji Ma, Wen-Cheng Wu, and Tsu-Chin Tsao are with the Department of Mechanical and Aerospace Engineering, University of California, Los Angeles, CA, USA.

{\tt\footnotesize \{yutingkevinlai, jacobrosen, yforoutani, jima, leox2x924, ttsao\}@ucla.edu.}}%
\thanks{$^2$Jean-Pierre Hubschman is with the Stein Eye Institute, University of California, Los Angeles, CA, USA. 
{\tt\footnotesize hubschman@jsei.ucla.edu.}}}

\maketitle

\begin{abstract}

Despite the extensive demonstration of robotic systems for both cataract and vitreoretinal procedures, existing technologies or mechanisms still possess insufficient accuracy, precision, and degrees of freedom for instrument manipulation or potentially automated tool exchange during surgical procedures.
A new robotic system that focuses on improving tooltip accuracy, tracking performance, and smooth instrument exchange mechanism is therefore designed and manufactured.
Its tooltip accuracy, precision, and mechanical capability of maintaining small incision through remote center of motion were externally evaluated using an optical coherence tomography (OCT) system.
Through robot calibration and precise coordinate registration, the accuracy of tooltip positioning was measured to be 0.053$\pm$0.031 mm, and the overall performance was demonstrated on an OCT-guided automated cataract lens extraction procedure with deep learning-based pre-operative anatomical modeling and real-time supervision.

\end{abstract}


\section{Introduction} \label{introduction}

\noindent Surgical robots have demonstrated success in improving the safety and efficiency of surgical procedures in recent decades \cite{10.1093/jsprm/snac003}. 
Intraocular procedures often require high precision and delicate manipulation to reduce surgical complications, but some procedures remain infeasible to humans due to physiological limitations \cite{gerber2019optical}, left alone retinal vein cannulation that is too risky to be performed by human surgeons \cite{ourak2019combined}.
Even experienced surgeons include an average hand tremor of 200 to 350 \textmu m relative to vein diameters of 120 to 200 \textmu m, as well as the inability to perceive the small manipulating forces associated with piercing veins \cite{vander2020robotic, gupta1999surgical}.
As a result, surgical robots become good candidates for this type of sophisticated tissue manipulation.

\begin{figure}[t!]
\centering
\includegraphics[width=0.9\linewidth]{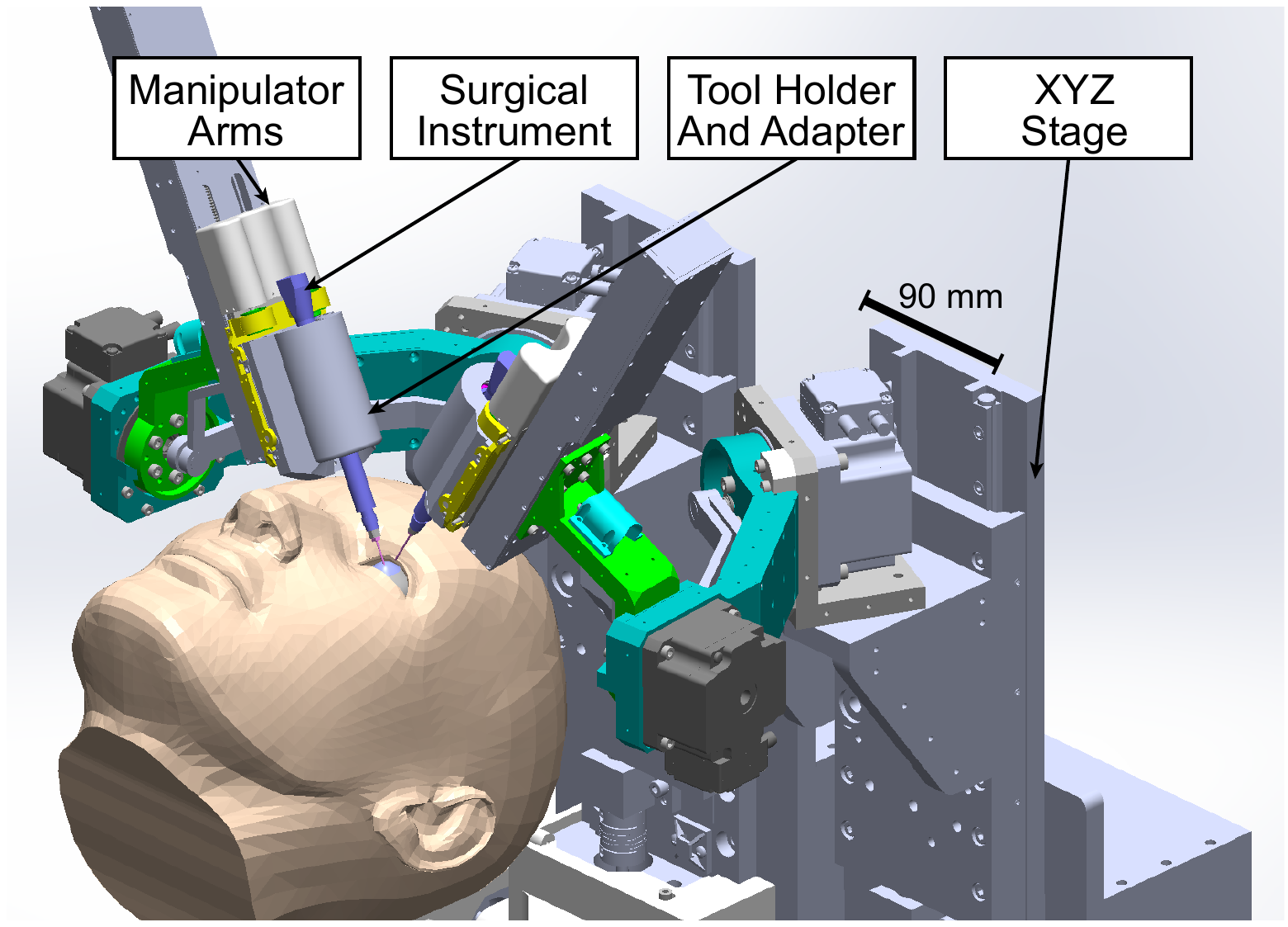}
\caption{The computer-aided design (CAD) model of the developed robotic system (isometric view) which illustrates the translational stage, two robotic arms, surgical instrument, and tool adapters and holders.}
\label{fig:IRISS2System}
\end{figure}

Computer Aided Design (CAD) and Computer Aided Manufacturing (CAM) were the sources of inspiration for the introduction of surgical robotic systems into the operating room. 
Automated robotic intervention divides surgeries into three main steps: (1) the pre-operative step in which an anatomical map was constructed with computer-assisted modeling, diagnosing, and planning, (2) the intraoperative step in which surgical plans are updated based on real-time sensory feedback, and (3) the post-operative step in which post-surgical assessment is conducted and stored in a repository {\cite{taylor2001computer}}, {\cite{taylor2016medical}}, {\cite{taylor2020computer}}. 

Three underlying assumptions of such intervention are: (1) the existence of real-time imaging modalities, (2) the ability to create model, perform diagnosis, and modify plans during the surgery, and (3) the surgical site is static (structured environment) or quasi-static between consecutive scans of the anatomy.
However, for surgical interventions involving soft tissues, these conditions present significant challenges because of their dynamic nature. 
As a result, early introduction of surgical robotic platforms for soft tissues utilized teleoperation modes which includes human surgeons {\cite{rosen2011surgical}} while avoiding automated strategies as the conditions listed above can not be met.
Two soft tissues organs, including the brain and eye, are unique among other human body soft tissues organs as they are encapsulated by skull or skull orbit. 
From a robotic perspective, this encapsulation allows one to treat these two organs as structured environments. 
This assumption is further reinforced for the eye since the pressures in different chambers of the eye tend to maintain the internal anatomical structures as long as a minimally invasive approach is used and the pressure levels can be regulated and maintained.

Minimally invasive surgical robotic systems have been developed to assist various intraocular surgical procedures.
The IRISS \cite{wilson2018intraocular, chen2018intraocular}, Preceyes \cite{meenink2013robot}, MYNUTIA \cite{gijbels2018human}, and Steady-Hand Eye Robot (SHER) \cite{taylor1999steady} are some examples of surgical robots that have been developed to be used by the surgeon for teleoperated and semi-automated procedures, which reduce hand tremors and increase surgical safety {\cite{gerber2020advanced}}, {\cite{ladha2023advantages}}.
The surgeon is indirectly in contact with the patient by manipulating a joystick which sends commands to the remote robot to operate the tool on the patient.
However, prolonged training process and insufficient maneuverability of such teleoperated robotic systems is tedious and time-consuming {\cite{sridhar2017training}}, making a high degrees-of-freedom (DOF) automated surgical robotic system critical to compensate for such limitation.


This work intends to address the current limitations by developing a new robotic system for intraocular manipulations with the focus on the details of mechanical design and the evaluation of the developed system.
The main highlights of the new system are as follows:
\begin{itemize}
    \item A total of five DOF mechanism and actuation that is capable of performing a wide range of surgical procedures, including tool rotation and injection.
    \item A lock-and-key mechanism with magnetic interface that allows smooth and repeatable tool exchange for different surgical instruments, and can be seamlessly integrated with an instrument magazine for automated tool exchange.
    \item The Denavit-Hartenberg (DH) parameters were optimized to increase robot reachable workspace without colliding with human patient while maintaining minimal link lengths and masses.
    \item Additional robot DH calibration for improved accuracy and precision and account for misalignment errors.
\end{itemize}
With these mechanical and software improvements, we hypothesize that the system possesses the accuracy and precision needed throughout cataract or retina surgery.
The developed system is integrated with real-time imaging from an Optical Coherence Tomography (OCT) system and a digital camera.
The performance of the tooltip was evaluated through the OCT and the surgical effectiveness of the system was experimentally assessed with an OCT-guided cataract lens extraction procedure on 30 \textit{ex-vivo} pig eyes.

\section{Mechanical Design} \label{method}

\subsection{Manipulator Design}

\begin{figure}[t!]
\centering
\includegraphics[width =\linewidth]{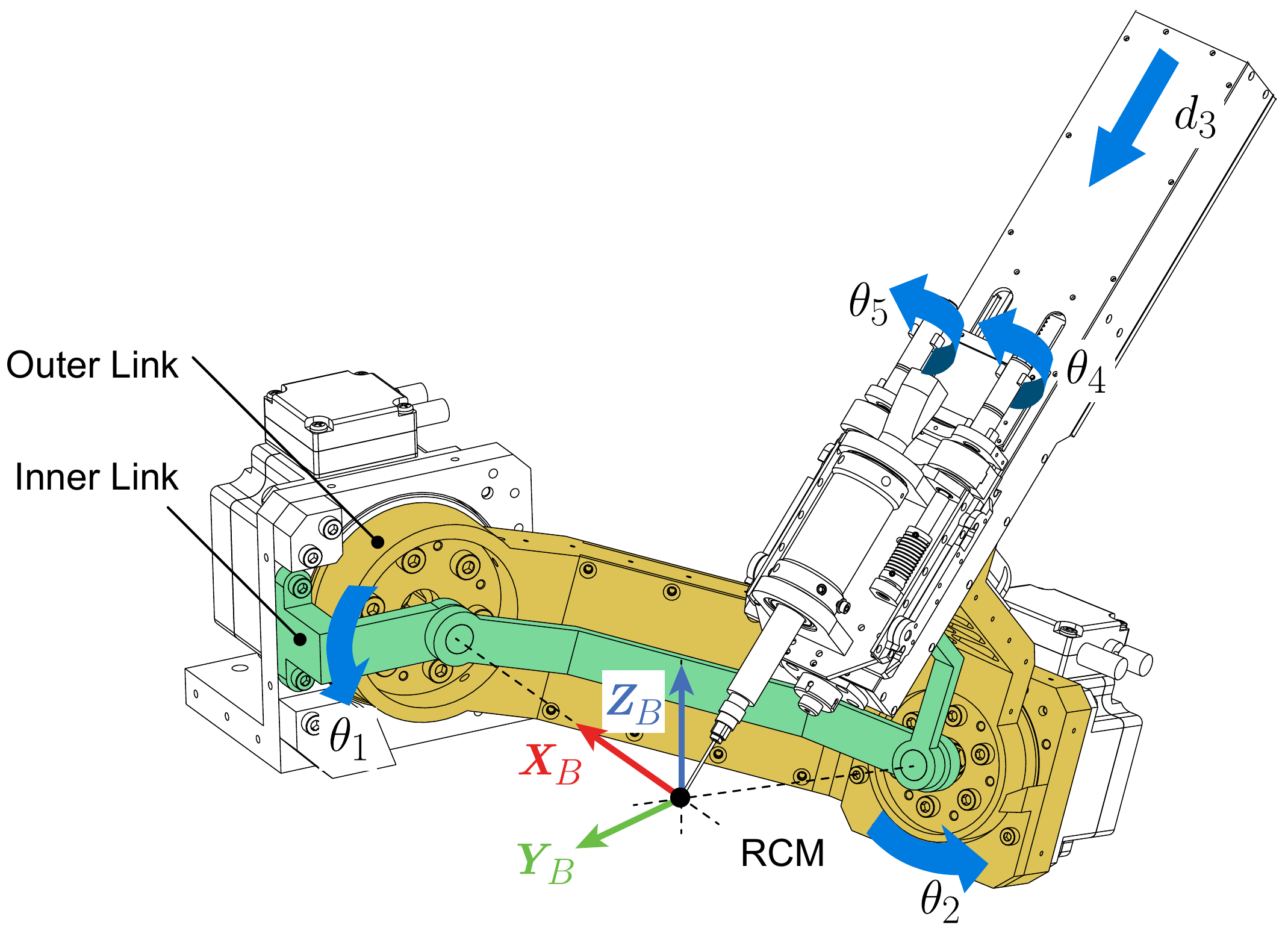}
\caption{Shown is the illustration of the robotic arm with joint definition and the robot coordinate frame. All axes of rotation and translation coincide at the RCM. The configuration shown is with $\theta_1=\theta_2=-10^\circ$, $\theta_4=0$, and $d_3=d_5=0$. Arrows indicate the positive direction of each joint.}
\label{fig:armAxisDefinition}
\end{figure}

\noindent The new robotic system contains two robotic arms that has the capability of performing bi-manual operation (Fig. \ref{fig:IRISS2System}).
The spherical parallel mechanism (SPM) is used in the design of each robotic arm, where the intersection of the axes of rotation coincide at the remote center of motion (RCM) to ensure minimal incision.
The first two links of each robotic arm are structured as parallel mechanisms to achieve high rigidity and vibration reduction of the robot.
The outer two thicker links (blue) were motorized and the inner two links (cyan) closer to the RCM are passive and were added to increase the overall stiffness of the mechanism.
Both sets of links share the same first two rotation axes (Fig. \ref{fig:armAxisDefinition}).

The first ($\theta_1$) and second ($\theta_2$) joints are two rotational joints separating 60$^\circ$ apart and their rotational axes are aligned to enforce the RCM.
The third joint ($d_3$) is the in-and-out translational motion of the surgical instrument where the centerline passes through the RCM.
Regardless of robotic arm motion, the RCM point remains fixed at the incision point of an eyeball.
In addition to the abovementioned three joints, additional two joints are located at the end of the arm which do not affect the end-effector position in the ideal case.
The fourth joint ($\theta_4$) is used to control the rotation of the entire surgical instrument about its centerline, while the fifth joint ($d_5$) is to control the gripper and injection motions for the use of forceps, syringes, or intraocular lens injectors.
Therefore, a single arm has a total of five degrees of freedom (DOF), and the joint limits are shown in Table {\ref{tab:jointLimits}}.
Each arm is mounted on a commercially available translational stage (Optic Focus MOXYZ-02-100-100-100) in all $x$, $y$, and $z$-direction.
The travel length for each axis is 100 mm, which can account for potential patient eyeball movement during the surgical procedure.

\begin{table}[t!]
    \centering
    \caption{The limits of each joints.}
    \begin{tabular}{c|c|c|c|c|c}
     & Joint 1 & Joint 2 & Joint 3 & Joint 4 & Joint 5 \\ \hline
    Max & 0$^\circ$ & 40$^\circ$ & 25 mm & 720$^\circ$ & 500 mm\\
    Min & -70$^\circ$ & -40$^\circ$ & -40 mm & -720$^\circ$ & 0 mm
    \end{tabular}
    \label{tab:jointLimits}
\end{table}

\begin{figure}[t!]
\centering
\includegraphics[width=0.85\linewidth]{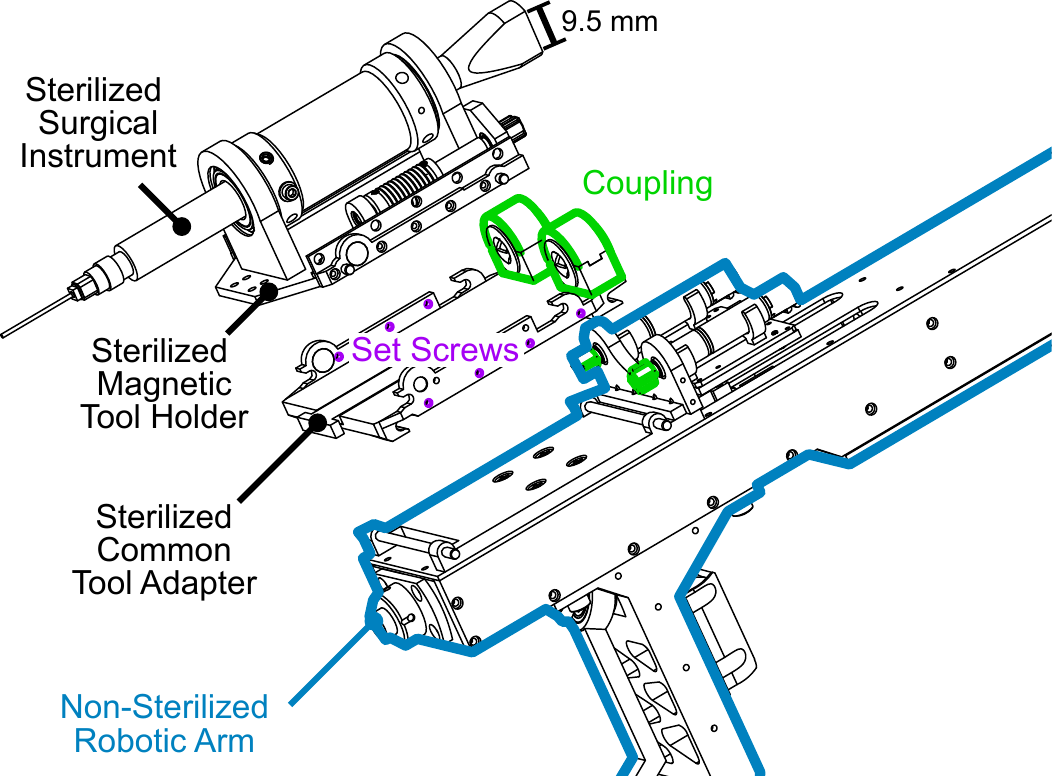}
\caption{Surgical tool interface on the developed robotic arm. The blue area indicates the non-sterilized robotic arm. The green area marks the mechanical coupling between joints 4 and 5 and the tool. The purple circles represent set screws that can align the tooltip $\hat{Z}$ axis with the joint 3 motion.}
\label{fig:roboticArmTool}
\end{figure}

To meet the sterilization requirements in the operating room, a tool adapter that can be sterilized is designed to serve as the separation between the non-sterilized robotic arm and the sterilized tool (Fig. {\ref{fig:roboticArmTool}}).
Before each procedure, only the magnetic tool holder and common tool adapter need to be changed and sterilized.
Although the tool holders are with various designs to accommodate different surgical procedures, the common tool adapter enables the mounting of these tool holders onto the arm easily.
The attachment between the tool holder and the adapter utilizes magnet with lock-and-key self-aligned features to minimize relative motion between the two parts, and the tool holder can be easily detached from the arm to facilitate fast tool exchange.
The transmission from the motors to the surgical instrument is realized with a mechanical coupling on the tool adapter without any direct wiring between the two.
To minimize the horizontal displacement between the z-axis of the tooltip frame and the joint 3 motion (d3), set screws were added on the common tool adapter to reduce the misalignment.

\subsection{DH Parameter Optimization}

\noindent To allow sufficiently large intraocular manipulation, DH parameter optimization on the link parameters was numerically computed in Matlab (Mathworks, Inc., Naticks, MA, USA) to find the best link configuration that can achieve the largest workspace.
Instead of using the Jacobian, which only provides information about the reach and manipulability of the robot, we introduce a performance score that considers reachable workspace, global conditioning index, endpoint stiffness, and link length.

\begin{equation}
\textnormal{Score} = \eta \frac{K_{end}}{Q_L \cdot \alpha^3},
\end{equation}
where $\eta$ is the Global Conditioning Index (GCI) in the cone workspace ($0<\eta<1$) {\cite{gosselin1991global}}, $K_{end}$ is the endpoint stiffness, which assumes the robot is rigid and calculates the overall deflection of the links in each robot link configuration {\cite{doria2019analysis}}.
$Q_L=L/V^{1/3}$ is the structural length index which calculates the ratio of the robot manipulator length sum to the cube root of the workspace volume {\cite{kucuk2006comparative}}, and $\alpha$ computes the arc length sum of all links except the base link, which is calculated by $\alpha=\sum^n_{i=2}(r_i\cdot a_{i-1})$.
If $\eta$ is closer to 1, it means the manipulator would have better performance score in all configurations, therefore it is to be maximized over the parameters.

To compute $K_{end}$, a mathematical stiffness model \cite{asada1991robot} is used to correlate the force ($F$) and the deflection ($\Delta x$):

\begin{equation}
    \Delta x=J(q)K_q^{-1}J(q)^T F
\end{equation}
where J(q) is the Jacobian matrix that depends on robot configuration, and $K_q$ is diagonal and can be obtained by solving the least-squares problem.
$K_{end}$ is computed to be the average of the minimal $K_q$ in the workspace, which is then normalized to be within 0 and 1.
Larger $K_{end}$ will result in larger score, which corresponds to larger robot stiffness and better position accuracy in the presence of disturbance forces.

\begin{figure}[t!]
\centering
\includegraphics[width=0.75\linewidth]{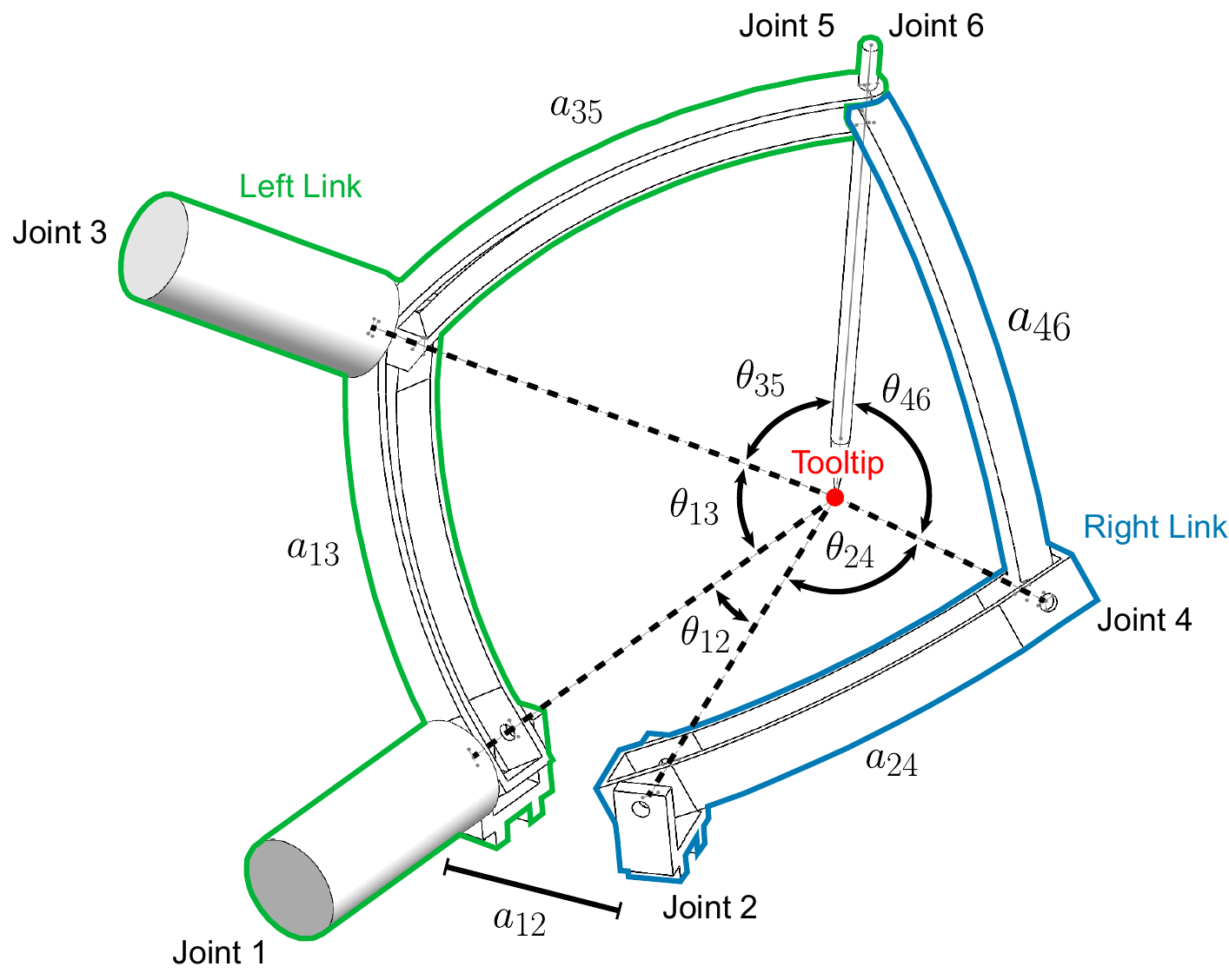}
\caption{Spherical parallel mechanism used for robotic arm optimization. The joint and link definition for the optimization are different from the actual robot design in Fig. \ref{fig:armAxisDefinition} and \ref{fig:roboticArmTool}.}
\label{fig:parallelSphericalDesign}
\end{figure}

To calculate $Q_L=L/V^{1/3}$, $L$ is the length sum of the each link which is calculated as $L=\sum^n_{i=1}(a_{i-1}+d_{i-1})$, and $a_{i-1}$ and $d_i$ are the link length and joint offset in mm. 
$V$ is the volume of reachable workspace in mm\textsuperscript{3}.
For an ideal SPM, the goal is to find a minimum $L$ that can achieve maximum $V$, which results in a smaller $Q_L$ and a larger overall score.
Finally, the $\alpha$ values are normalized to the range of 0 and 1, which means each link is 0$^\circ$ and 90$^\circ$ apart.
It is ideal to have minimal $\alpha$ for less structure mass and more structure rigidity.
In addition, to use minimal robot link lengths to cover the head without collision, the biggest and smallest sizes of head are considered into the design.
However, different sizes of noses or eyes were not taken into consideration because the robot is not able to collide with those areas.

\begin{figure}[!h]
\centering
\includegraphics[width =\linewidth]{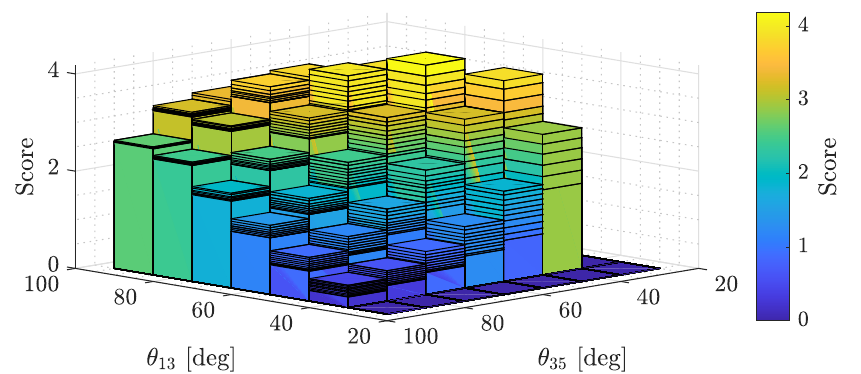}
\caption{Arm optimization result for different angle configurations.}
\label{fig:arm_optimization_result}
\end{figure}

Instead of optimizing the whole sphere workspace, we used a semi-sphere and divided it into 2592 small grids, each containing a 5$^\circ\times$5$^\circ$ area.
The optimization procedure iterates through these grids and calculates the best scores, and finally chooses best link parameters from all results.
A symmetric structure that contains the left link and the right link (Fig. \ref{fig:parallelSphericalDesign}) that are $a_{12}$ apart is considered to account for possible singularity.
We assume $a_{13}=a_{24}$ and $a_{35}=a_{46}$, and each link contains three joints.
$a_{13}$ and $a_{35}$ are connected with serial passive rotational joints from the base to the tooltip.
The angle between each joint is defined as $\theta_{ij}$, including the angle between the bases of the link ($\theta_{12}$), where $i$ and $j$ are the index of the joint.
The optimization process then calculates the score based on these parameters.
Fig. \ref{fig:arm_optimization_result} shows the map of the scores from the optimization and the optimal score is achieved when both $\theta_{13}$ and {$\theta_{35}$} are 60$^\circ$.
Also from the results $\theta_{12}=0^\circ$, which means the base of the two links coincide.
Hence, a single link is able to reach all the points in the workspace without any singularity.

\subsection{Manipulator Kinematics} \label{sec:manipulator_kinematics}

\noindent The first three degrees of freedom ($\theta_1$, $\theta_2$, $d_3$) in the robotic arm naturally establish the RCM.
Ideally, we assume the rotation of the surgical instrument ($\theta_4$) and its injection motion ($\theta_5$) do not affect the RCM, so they are ignored in the kinematics formulation.
The equations for the robot forward kinematics are presented below ($g_{st}(\theta)$).

\begin{equation}
\begin{aligned}
    \begingroup
    \renewcommand*{\arraystretch}{1.5}
    \begin{bmatrix} [ccc|c]
    c_1c_2-\frac{1}{2}s_1s_2 & -\frac{1}{4}(3s_1-2c_1s_2-c_2s_1) & \eta_1 & d_3\eta_1
    \\
    \frac{1}{2}c_1s_2-c_2s_1 & \frac{1}{4}(3c_1+c_1c_2 -2s_1s_2) & \eta_2 & d_3\eta_2 \\
    -\frac{\sqrt{3}}{2}s_2 & \frac{\sqrt{3}}{4}(1-c_2) & \eta_3 & d_3\eta_3  \\ \hline
    0 & 0 & 0 & 1 
    \end{bmatrix} 
    \endgroup 
    \\
\end{aligned}
\end{equation}
with $\eta_1 = (2\sqrt{3}c_1s_2+c_2s_1-\sqrt{3}s_1)/4$, $\eta_2 = \sqrt{3}(c_1-c_1c_2-2s_1s_2)/4$, $\eta_3 = (3c_2 + 1)/4$. Here $c_i = \textnormal{cos}(\theta_i)$, $s_i = \textnormal{sin}(\theta_i)$.

The inverse kinematics is calculated by the following equations given a target tooltip position $\textbf{*p}=[p_x \; p_y \; p_z]^T$ within the workspace of the robot. 

\begin{equation}
\begin{aligned}
    d_3 &= \sqrt{p_x^2 + p_y^2 + p_z^2} \\
    \theta_2 &= \textnormal{asin}(\xi_1,\xi_2) \\
    \theta_1 &= \textnormal{atan}(\xi_3,\xi_4)
\end{aligned}
\end{equation}
, where

\begin{equation}
    \begin{aligned}
        \xi_1 &= 4p_z-d_3 \\
        \xi_2 &= 3d_3 \\
        \xi_3 &= \frac{\sqrt{3}}{2}p_x+\sqrt{3}p_ycos(\theta_2)-\frac{\sqrt{3}}{2}p_xsin(\theta_2) \\
        \xi_4 &= \sqrt{3}p_xcos(\theta_2)-\frac{\sqrt{3}}{2}p_y+\frac{\sqrt{3}}{2}p_ysin(\theta_2).
    \end{aligned}
\end{equation}

Potential misalignment due to assembly errors will stack-up from the robot base frame to the end-effector and affect the accuracy of the surgical tooltip.
Rotational variations about the $\hat{Z}$ and $\hat{Y}$ axes can be ignore because these can be compensated by the motions of $\theta_1$ and $\theta_2$.
Variation about the $\hat{X}$ axis presents, but it is considered negligible because the SPM is manufactured by high-accuracy and repeatable CNC machines. 
Therefore, the major source of assembly error results from the assembly error between the instrument adapter and the surgical tool holder.
This error, can be minimized by applying DH calibration method, which will be covered in Section {\ref{sec:evaluation_and_calibration}}.
As a result, a numerical solver is needed for solving inverse kinematics of the robot instead of a standard analytical solution.

\section{Software Design and Mechatronics} \label{sec:mechatronifcs}

\subsection{Software Design}
\begin{figure}[t!]
    \centering
    \includegraphics[width =0.9\linewidth]{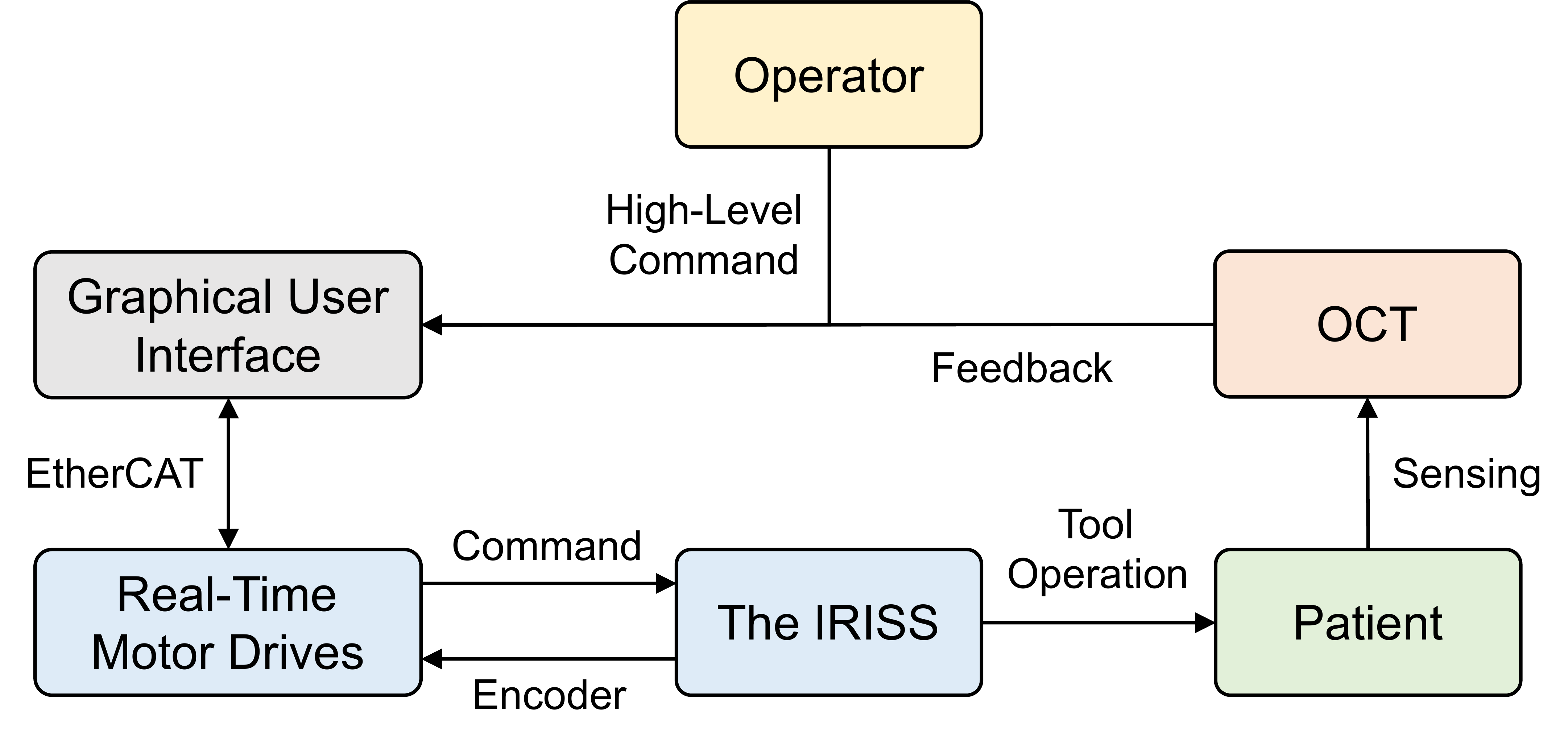}
    \caption{System architecture}
    \label{fig:system_architecture}
\end{figure}

\noindent The system architecture is shown in Fig. \ref{fig:system_architecture}.
The robotic system incorporated the optical coherence tomography (OCT) system (Telesto II 1060LR with objective lens LSM04BB; ThorLabs).
This enables three-dimensional acquisition of the eye anatomy, followed by image processing techniques to obtain tissue locations.
The operator can give high-level commands to the graphical used interface for trajectory generation based on the identified tissue locations.
Then the trajectory serially communicates to real-time motor drives through EtherCAT with a sampling frequency of 1000 Hz and a latency less than 1 ms.
During the motion of the surgical tool, the reference commands for joint angles are generated and the positions are measured by rotational optical encoders, with the manufacturer-reported resolutions shown in Table \ref{tab:motor_info}.

\begin{table}[t!]
\caption{The resolution of each motor in the robot}
\begin{center}
\begin{tabular}{c| c| c| c}
Joint & Count/Rev. & Gear Ratio & Resolution\\
\hline
1 & 8000 & 100:1 & 0.0054 deg/count \\
2 & 8000 & 100:1 & 0.0054 deg/count \\
3 & 1024 & 16:1 & 0.2550 \textmu m/count \\
4 & 256 & 15:1 & 0.0230 deg/count \\
5 & 256 & 15:1 & 0.0230 deg/count \\

\end{tabular}
\label{tab:motor_info}
\end{center}
\end{table}

\subsection{Controller Design}
\begin{figure}[t!]
    \centering
    \includegraphics[width=0.8\linewidth]{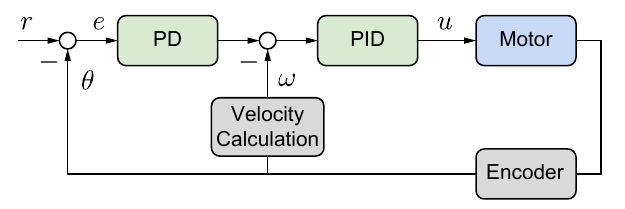}
    \caption{Illustration of the real-time control block diagram for each motor.}
    \label{fig:control_block_diagram}
\end{figure}

\noindent The robot is actuated by eight brushed DC servo motors, including stage motion that moves the entire robotic arm in the $x$, $y$, and $z$-direction.
The controllers are implemented on the motor drives and the architecture is shown in Fig. \ref{fig:control_block_diagram}.
The gear ratio and output resolution for each motor are shown in Table \ref{tab:motor_info}
The first two joints ($\theta_1, \theta_2$) are actuated by Harmonic drives FHA-11C and FHA-8C motors. 
The third joint ($d_3$) is actuated by Maxon motor DCX10L that drives a miniature lead screw with a pitch of 1.219 mm/turn.
The load capacity of $d_3$ was designed to be 0.9 kg with a maximum allowable linear speed of 15.36 mm/s, which enables fast instrument retraction when emergency occurs during intraocular operation.
To minimize the position measurement error from encoder motor shaft to the linear output, a miniature magnetic linear sensor (RLC2HD; RLS Merilna tehnika d.o.o.) was mounted directly on the linear rail.
Two miniature Maxon DCX6M motors were used to drive the tool rotation and injection motion.
The spindle on the gear shaft directly drives the tool adapter, which then transmits torques to two tool rotary inputs. 
One rotary input is to drive tool shaft rotation though capstan transmission mechanism, while the other rotary input is to drive tool functions such as grasping or injection though lead screw transmission mechanism.

The precision of each joint was characterized by performing incremental motion with the output measured by a digital gauging probe (MAGNESCALE LY-201, Sony) that has a resolution of 1 \textmu m.
Since mechanical design of the robot contains an RCM, the probe indirectly measured the motion of joint 1 and 2 at the outer links with an equivalent incremental motion of 1 \textmu m, whereas direct linear incremental motion was applied on joint 3.
These incremental joint motions result in less than 1 \textmu m precision and a 1 \textmu m backlash at the measuring location at the link, which is equivalent to 0.2 \textmu m at the tooltip.
For the stage movement, similar measurement was conducted and the accuracy was less than 5 \textmu m with a precision of 1 \textmu m.
Together, the resulting tooltip accuracy from incremental joint motions was approximately 0.25 \textmu m.

\begin{figure}[t!]
    \centering
    \includegraphics[width=0.85\linewidth]{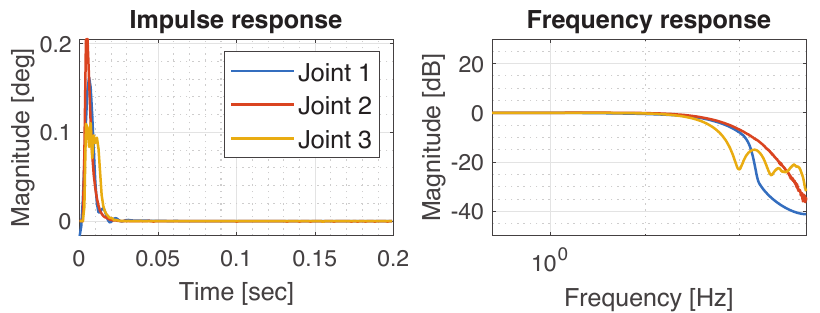}
    \caption{Impulse response and frequency response of each joint}
    \label{fig:robot_bandwidth}
\end{figure}

\subsection{Manipulator Dynamics}

\noindent Closed-loop system identification was done with a unit step reference signal applied to each joint. 
To obtain the frequency response, fast fourier transform was performed on the output impulse response calculated from the step response (Fig. \ref{fig:robot_bandwidth}).
Table {\ref{tab:robot_bandwidth}} shows that the tracking bandwidth of each joint is higher than 13 Hz, due to the compact and lightweight design of the robot and motor selection.
Therefore, the motors are able to track higher frequency commands such as a faster trajectory or performing an fast emergency retraction.
Only the first three joints are compared because joint 4 and joint 5 are rotation and grasping motion which do not affect the tooltip tracking performance.
In addition, no vibration mode is observed from the frequency response of each joint, which means the risk of exciting resonance modes is minimized while the robot is tracking a trajectory.

\begin{table}[t!]
\caption{Motors tracking bandwidth in the robot}
\begin{center}
\begin{tabular}{c | c}
Joint & Bandwidth [Hz] \\ \hline
$\theta_2$ & 37.5 \\
$\theta_2$ & 18  \\ 
$d_3$ & 13 \\ 
\end{tabular}
\label{tab:robot_bandwidth}
\end{center}
\end{table}



\section{System Evaluation and Performance}\label{sec:evaluation_and_calibration}

\noindent This section uses the OCT as the coordinate measurement machine for tooltip localization and presents the accuracy and precision of the robot, as well as describing the procedure of DH parameter calibration for accurate tooltip positioning.
The OCT volume scan is a good candidate for tooltip localization because it constructs a point cloud data and has an axial resolution of 9.2 \textmu m and a lateral resolution of 25 \textmu m.

\subsection{Tooltip Localization}
\begin{figure}
    \centering
    \includegraphics[width=0.8\linewidth]{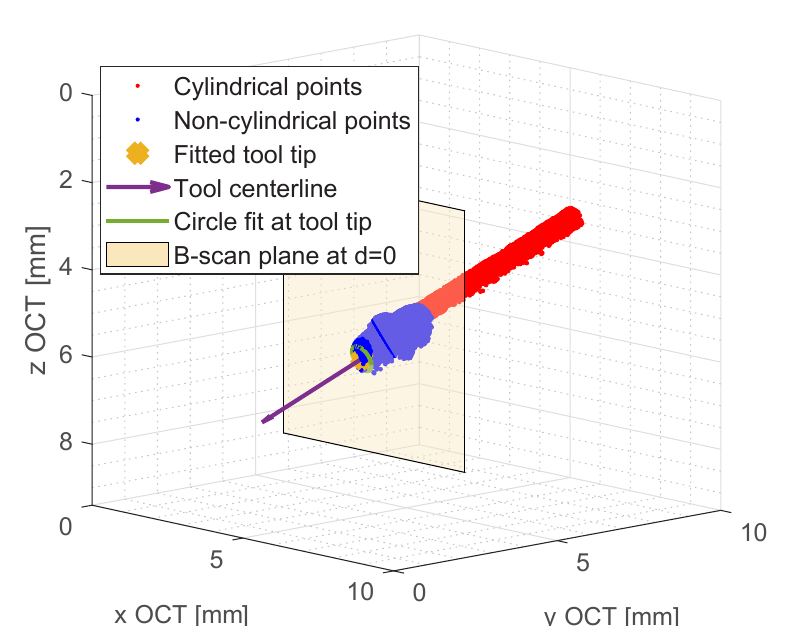}
    \caption{Tool localization method with color-coded illustration. The yellow plane indicates $d=0$ in the cumulative percentage calculation for tooltip localization.}
    \label{fig:tool_localization}
\end{figure}

\noindent To evaluate the performance of the robotic arm, a tool pose detection method was developed using OCT volume scans.
After each OCT acquisition, thresholding was performed to separate the point cloud of the tool from the background.
The points with higher intensities were considered as the tool and were reconstructed in the three-dimensional space as the tool point cloud (Fig. \ref{fig:tool_localization}).
To account for possible non-cylindrical features at the tooltip, the points pertaining to such features were automatically discarded based on an empirical distance threshold value derived from all volume scans.
Principal component analysis was then performed on the overall tool point cloud, followed by a Gauss-Newton method \cite{gauss_gauss_1809} to obtain a refined tool orientation (purple line).
Since the tool shaft is cylindrical, the point cloud obtained from the OCT can only partially determine the tool orientation and the $\hat{Z}$-axis of $\{t\}$ in Fig. \ref{fig:armAxisDefinition}.
Hence, the direction of the $\hat{X}$ and $\hat{Y}$ axes remains undetermined from a single point cloud.


\begin{figure}
    \centering
    \includegraphics[width=0.9\linewidth]{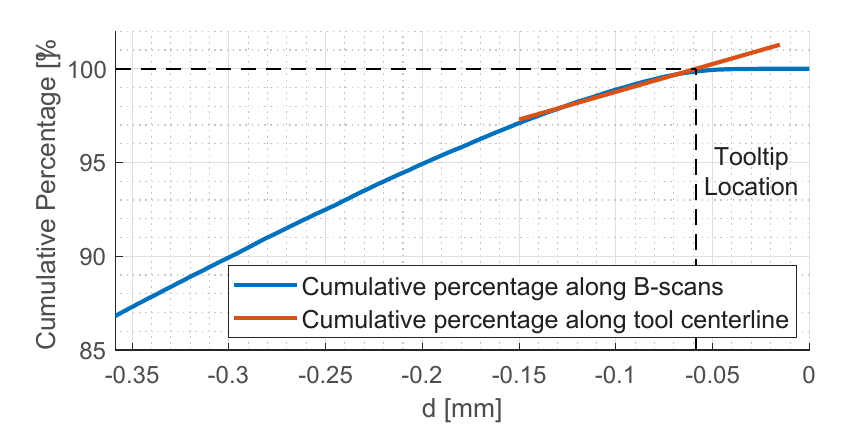}
    \caption{Cumulative percentage of point cloud vs $d$, where $d$ represents the location of each point in the point cloud along the B-scan direction and fitted tool centerline.}
    \label{fig:cumulative_plot}
\end{figure}

The actual tooltip position cannot be adequately estimated because the tool centerline does not necessarily align with the B-scan direction.
Hence a modified tooltip estimation was developed which uses the linear interpolation of the cumulative percentage curve of the tool point cloud constructed between (a) along the B-scan direction, and (b) along the fitted tool centerline.
The cumulative percentage sums up to 100\% at the tooltip, where $d=0$ is defined (Fig. {\ref{fig:cumulative_plot}}).
Note that only the points within the 0.15 mm range from the tooltip are considered to ensure accurate interpolation.
After the interpolation, the actual tooltip location can be determined with the $d$ from the tooltip along the tool centerline.
The error of the localization method was identified (Table \ref{tab:octPrecisionError}) by performing 32 repeating OCT measurements on an I/A handpiece (92-IA21 Handle, Millennium Surgical). 

\begin{table}
\caption{Error of the localization method}
\begin{center}
{\setlength{\extrarowheight}{2pt}
\begin{tabular}{c|c|c|c}
error type        & rms   & max    & std      \\ [2pt]
\hline
position [mm]     & 0.015 & 0.030  & 0.008   \\ [2pt]
\hline
orientation [deg] & 0.035 & 0.086  & 0.014   \\
\end{tabular}}
\label{tab:octPrecisionError}
\end{center}
\end{table}

\subsection{Accuracy Evaluation and Robot Calibration}
\noindent Tooltip accuracy is evaluated by the position error ($e_p$) and orientation error ($e_z$) in the OCT frame ($\{m\}$), which are defined as the difference between $i$\textendash th OCT measurement and the estimated value from Forward Kinematics (FK) of the robot (\ref{eq:errorDefinition}).
To compare the two in the same coordinate system, coordinate transformation is required to transform the tooltip points from the robot base frame ($\{b\}$) to the OCT frame.

\begin{equation} \label{eq:errorDefinition}
\begin{aligned}
    e_i(\eta,t_{mb}) &\coloneqq 
    \begin{bmatrix}
        e_p\\
        we_z
    \end{bmatrix}_i\\
     &= 
     \begin{bmatrix}
        p_{m,i} - [\prescript{m}{b}{R}(r_{mb})p_{b,i}(q_i,\eta)+p_{mb}] \\
        w[z_{m,i}-\prescript{m}{b}{R}(r_{mb})z_{b,i}(q_i,\eta)],
    \end{bmatrix}
\end{aligned}
\end{equation}
and the error vector of all measurements is defined as
\begin{equation}\label{eq:errorVector}
    e(\eta,t_{mb}) \coloneqq 
    \begin{bmatrix}
        e_1^T & e_2^T & ... & e_n^T
    \end{bmatrix}^T,
\end{equation}
where $p_{m,i}$, and $z_{m,i}$ represent the tooltip position and orientation identified from tooltip localization. 
$w$ is a predetermined weighting between $e_p$ and $e_z$. 
$p_{b,i}$ and $z_{b,i}$ are the tooltip position and orientation estimated from FK parameters ($\eta$) based on encoder measurements ($q_i$).
$\prescript{m}{b}{R}$, $r_{mb}$, and $p_{mb}$ are Coordinate Transform (CT) parameters that characterize the relation between the OCT frame ($\{m\}$) and the robot base frame ($\{b\}$).
$\prescript{m}{b}{R}$ represents the rotation matrix from $\{m\}$ to $\{b\}$, whereas $r_{mb}\in R^3$ is its ZYX Euler angle representation.
$p_{mb} \in R^3$ is the location in $\{b\}$ relative to $\{m\}$.
The lumped CT parameters is defined as $t_{mb} \coloneqq \begin{bmatrix} p_{mb}^T & r_{mb}^T \end{bmatrix}^T \in R^6$.

Due to the existence of the parameter deviation, DH parameter calibration is required to reach the accuracy for surgical operations \cite{ouyang2015retinal}.
This means the calibrated FK model does not have the simple form discussed in Section {\ref{sec:manipulator_kinematics}}.
The FK estimation of $p_{b,i}$ and $z_{b,i}$ is derived from the following four-parameter homogeneous transform from the base frame $\{b\}$ to the tooltip frame $\{t\}$

\begin{equation}
\begin{aligned}
    \prescript{b}{t}{T}
    &= \begin{bmatrix}
        x_{b} & y_{b} & z_{b} & p_{b} \\
        0 & 0 & 0 & 1        
      \end{bmatrix} \\
    &= T_1 T_2 T_3 T_4
\end{aligned}
\end{equation}
,where $T_j$ represents the homogeneous transform from the current joint frame $\{j-1\}$ to the next joint frame $\{j\}$. 
For $j=[1,2,3]$, $T_j$ is formulated using the standard DH representation, where $\eta_j = [d_j, \theta_j, a_j, \alpha_j]$.

\begin{equation}
    T_j(\eta_j) = \textit{Trans}_z(d_j)\textit{Rot}_z(\theta_j)\textit{Trans}_x(a_j)\textit{Rot}_x(\alpha_j).
\end{equation}

The standard four-parameter representation has insufficient parameters to represent arbitrary tooltip position and tool orientation \cite{ZHUANG1993287,125952}, thus the following six-parameter representation is used for the last joint $\eta_4 = [d_4, \theta_4, a_4, b_4, \beta_4, \alpha_4]$.

\begin{multline}
    T_4(\eta_4) = \textit{Trans}_z(d_4)\textit{Rot}_z(\theta_4)\textit{Trans}_y(b_4)\textit{Rot}_y(\beta_4)\\ \textit{Trans}_x(a_4)\textit{Rot}_x(\alpha_4)
\end{multline}
The robot calibration optimizes both CT+FK parameters and is defined as $\gamma \coloneqq [\eta^T\:  t_{mb}^T]^T$.
To perform calibration, the robot was commanded to 30 randomly generated poses that were sufficiently large enough to cover the robot workspace. 
The least square objective function can be defined as:

\begin{equation}\label{eq:gammaDefinition}
    S(\gamma) \coloneqq e^Te = \sum_{i=1}^{n} e_i^T e_i
\end{equation}

\begin{equation}\label{eq:optimizationProblem}
    \gamma^* = \textit{arg} \: \min_{\gamma} S(\gamma)   
\end{equation}
Levenberg\textendash Marquardt (LM) optimization procedure \cite{10.2307/43633451, doi:10.1137/0111030} was conducted to solve for the optimal model parameters $\gamma^*$. 



Some parameters are fixed due to the observability of the parameters, which can be determined using Singular Value Decomposition analysis on the Jacobian matrix \cite{4791138,doi:10.1177/027836499101000106,9512730}.
When single or multiple parameter deviations cause little to no combined movement on the tooltip position and tool orientation, such parameter deviations can hardly be observed by measurement since they contribute little or no difference to the objective function (\ref{eq:gammaDefinition}).
This can be solved by either choosing another model parameterization or fixing some of the unobservable parameters at the nominal value and identifying the rest.


\begin{table}[t!]
\caption{Tooltip accuracy}
\begin{center}
{\setlength{\extrarowheight}{1pt}
\begin{tabular}{c| c |c |c| c}
error type  & parameter & RMS   & Max    & Std      \\ [1pt]
\hline
position    & CT        & 0.479/0.454 & 0.984/0.913  & 0.236/0.154   \\[1pt]
[mm]        & CT + FK   & 0.053/0.056 & 0.150/0.169  & 0.031/0.034 \\ [1pt]
\hline
orientation & CT        & 2.78/2.71 & 3.81/4.02  & 0.61/0.69   \\[1pt]
[deg]       & CT + FK   & 0.23/0.23 & 0.58/0.57  & 0.14/0.12 \\[1pt]
\multicolumn{5}{l}{$^{\mathrm{a}}$reported as ``calibration error / validation points error''}
\end{tabular}}
\label{tab:accuracyError}
\end{center}
\end{table}

\begin{table}[t!]
\caption{Tooltip accuracy}
\begin{center}
{\setlength{\extrarowheight}{1.2pt}
\begin{tabular}{c| c |c |c| c}
Error Type  & Parameter & RMS   & Max    & Std      \\ [1pt]
\hline
Position    & Without Calibration        & 0.454 & 0.913  & 0.154   \\[1pt]
[mm]        & With Calibration   & 0.056 & 0.169  & 0.034 \\ [1pt]
\hline
Orientation & Without Calibration        & 2.71 & 4.02  & 0.69   \\[1pt]
[deg]       & With Calibration   & 0.23 & 0.57  & 0.12 \\[1pt]
\end{tabular}}
\label{tab:accuracyError}
\end{center}
\end{table}

Additional 30 validation points were used to verify the calibrated parameters.
Both calibration and validation errors were reported in Table \ref{tab:accuracyError} and separated by "f/".
To demonstrate the effectiveness of the calibration algorithm, four different surgical instruments were analyzed.
Note that the one without FK parameter calibration is also compared, where the parameters to be optimized only contains the CT parameters: $\gamma \coloneqq t_{mb}$.
It can be seen that the RMS error is reduced if the CT and FK calibration are conducted together, and it still holds true in the validation set.
The CT+FK calibration jointly considers the CT and FK hence the robot-to-OCT registration error and the counterweight effect are included.

\begin{figure}[t!]
    \centering
    \includegraphics[width=0.8\linewidth]{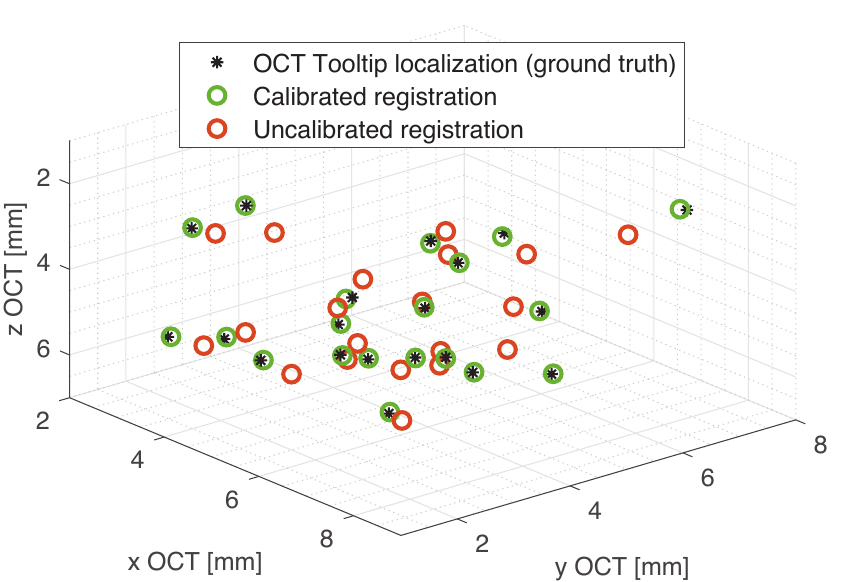}
    \caption{Comparison of registration results with and without calibration. Calibrated transformation includes CT+FK calibration. Each blue point is the tooltip location determined by the tooltip localization method.}
    \label{fig:compareRegistration}
\end{figure}

There is only a one-time calibration for each tool offline, and there is no need to re-calibrate the tool, assuming the robot assembly is rigid, no screw is loose, and the tooltip is not bent.
However, partial tool offset re-calibration procedure is required when there is tool exchange. 
This requires only a minimum of 5 OCT volume scans, and the total processing time is within 30 seconds.
This calibration process makes the inverse kinematics non-trivial, hence instead of using the analytical equations, a numerical solver is needed to accurately compute the joint angles given the tooltip positions.
The final CT+FK result is shown in Fig. \ref{fig:compareRegistration}, where the robot tooltip positions were transformed onto the OCT frame using the calibrated transformation and the uncalibrated transformation.
As shown in Table \ref{tab:CRError}, the root-mean-square (RMS) error of tooltip position and orientation is greatly reduced by 93\% if consider CT+FK parameter calibration.
This ensures the relationship between the two coordinate systems are well established and the projection error between the two are minimized.

\begin{table}[t!]
\caption{RMS Error comparison}
\begin{center}
{\setlength{\extrarowheight}{1.5pt}
\begin{tabular}{c|c|c}
Error Type  & Position [mm]  & Orientation [$^\circ$] \\
\hline
Calibrated Transformation & 0.038 & 0.137 \\
\hline
Uncalibrated Transformation  & 0.534 & 1.768  \\
\end{tabular}}
\label{tab:CRError}
\end{center}
\end{table}

\subsection{Precision Evaluation}

\noindent The precision is quantified by two different metrics: the tool exchange error and the tooltip repeatability.
The tool exchange error gauges the capability of the robot to maintain the tooltip location after multiple tool exchanges. 
The tool cartridge was repeatedly removed and mounted back on for 17 times while the robot remained at a fixed pose.
The results are reported in Table \ref{tab:toolExchangeError} with different tool types.
Robot repeatability evaluates the capability to move to the same pose after the same joint command is given. 
The robot was commanded to 5 different poses for 15 times with the tooltip position and tool orientation calculated from OCT measurements.
These 5 poses were chosen to be 2.5 to 4 mm apart to provide large pose variance while remaining in the robot workspace.
The repeatability error is reported in Table \ref{tab:repeatabilityError} and is defined as the standard deviation of the tooltip position or tool orientation of each pose. 
The repeatability error is consistent across different tools.

\begin{table}
\caption{tool exchange error}
\begin{center}
{\setlength{\extrarowheight}{1pt}
\begin{tabular}{c| c| p{1cm}| p{1cm} |p{1cm}}

error type  & tool              & rms   & max    & std      \\ [1pt]
\hline
            & cannula           & 0.083 & 0.159  & 0.041    \\ [1pt] 
position    & forceps           & 0.089 & 0.156  & 0.039    \\ [1pt]
[mm]        & vitreous cutter   & 0.251 & 0.512  & 0.129    \\ [1pt]
            & I/A               & 0.109 & 0.175  & 0.038    \\ [1pt]
\hline
            & cannula           & 0.045 & 0.080  & 0.019    \\ [1pt]
orientation & forceps           & 0.112 & 0.208  & 0.050    \\ [1pt]
[deg]       & vitreous cutter   & 0.148 & 0.271  & 0.071    \\ [1pt]
            & I/A               & 0.339 & 0.801  & 0.159    \\ [1pt]
\end{tabular}}
\label{tab:toolExchangeError}
\end{center}
\end{table}

\begin{table}
\caption{Tooltip repeatability error}
\begin{center}
{\setlength{\extrarowheight}{1pt}
\begin{tabular}{c| c| p{1cm}| p{1cm}| p{1cm}}
            & tool              & rms   & max    & std      \\ [1pt]
\hline
            & cannula           & 0.017 & 0.072  & 0.010    \\ [1pt] 
position    & forceps           & 0.011 & 0.025  & 0.005    \\ [1pt]
[mm]        & vitreous cutter   & 0.012 & 0.030  & 0.006    \\ [1pt]
            & I/A               & 0.019 & 0.071  & 0.014    \\ [1pt]
\hline
            & cannula           & 0.057 & 0.179  & 0.037    \\ [1pt]
orientation & forceps           & 0.043 & 0.092  & 0.021    \\ [1pt]
[deg]       & vitreous cutter   & 0.044 & 0.145  & 0.026    \\ [1pt]
            & I/A               & 0.047 & 0.121  & 0.025    \\ [1pt]
\end{tabular}}
\label{tab:repeatabilityError}
\end{center}
\end{table}

\begin{table}
\caption{RCM error in the Developed Robot}
\begin{center}
{\setlength{\extrarowheight}{1pt}
\begin{tabular}{c| c| c| c}

Type   & rms   & max    & std      \\ [1pt]
\hline
measured  RCM error [mm] & 0.14 & 0.23  & 0.05 \\ [1pt]
\hline
estimated RCM error [mm] & 0.13 & 0.19  & 0.05 \\ [1pt]
\end{tabular}}
\label{tab:RCM error}
\end{center}
\end{table}

\subsection{RCM Evaluation}
\noindent A new set of 30 different poses were selected to evaluate the ability to maintain the RCM.
RCM error of every measured tool pose can be calculated by solving (\ref{eq:rcmOptimizationProblem}) after computing the RCM error for each tool pose (\ref{eq:rcmError}).

\begin{equation}\label{eq:rcmError}
    e_{rcm,i} \coloneqq (I-z_{m,i}z_{m,i}^T)(p_{m,i}-p_{rcm})
\end{equation}

\begin{equation} \label{eq:rcmOptimizationProblem}
    p_{rcm}^* = arg \min_{p_{rcm}} \sum_{i=1}^{n} e_{rcm,i}^Te_{rcm,i}
\end{equation}
The new system has a RMS error of 0.14 mm that minimizes incision stress during surgical operation.
The effectiveness of CT+FK calibration can also be validated by calculating the estimated RCM based on the calibrated parameters $\gamma^*$.
The tool centerline measurement ($z_{m,i}$,$p_{m,i}$) in (\ref{eq:rcmError}) are replaced by estimated tool centerline with $\gamma^*$ ($z_{b,i}$,$p_{b,i}$). 
$p_{rcm}^*$ can then be calculated from the estimated RCM error using (\ref{eq:rcmOptimizationProblem}). 
The result for both the measured RCM error and estimated RCM error are both tabulated in Table \ref{tab:RCM error}.


\section{\textit{Ex-Vivo} Animal Study\\for Subretinal Injection}

\begin{figure}[t!]
\centering
\includegraphics[width=0.95\linewidth]{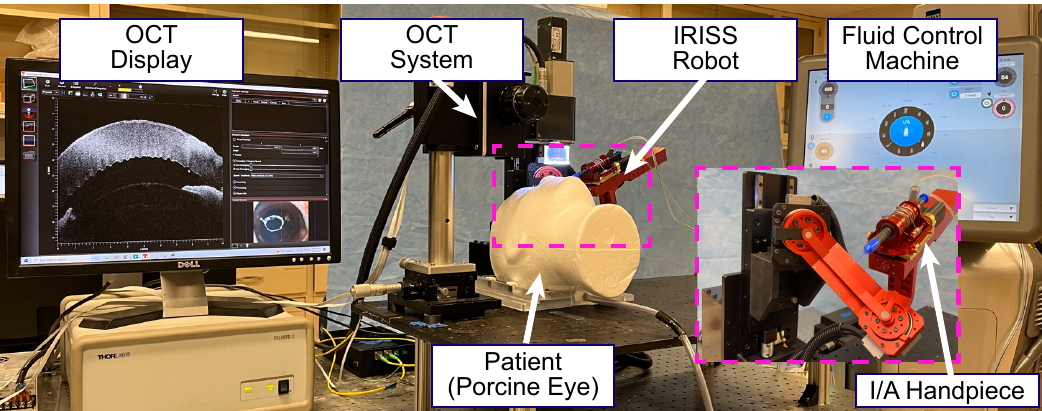}
\caption{Experimental setup for surgical tests on porcine eyes.}
\label{fig:setup}
\end{figure}



\subsection{Surgical Challenge}
Subretinal injection (SRI) remains one of the most challenging procedures in vitreoretinal surgery, requiring micron-level precision to deliver therapeutic agents into the subretinal space without damaging delicate retinal structures. 
The retinal pigment epithelium (RPE), the target layer for many of these therapies, lies beneath the internal limiting membrane (ILM) and is typically separated from the retinal surface by only  approximately 10 µm. 
Any overshoot or misalignment during tool insertion risks perforating the retina or damaging the RPE.
Traditional manual techniques rely on the stability of surgeon's hand and visual cues from the microscope images, which are insufficient for such delicate membrane.
Our developed robotic arm is specifically designed to address these challenges. 
With a remote center of motion (RCM) architecture optimized for intraocular access and an actuation system capable of a 1 \textmu m incremental motion resolution, the platform enables high-precision control required for safe and consistent subretinal delivery.

\subsection{Experimental Setup}

To evaluate the performance of the robotic system for subretinal injections, we conducted a series of \textit{ex-vivo} experiments using freshly enucleated pig eyes. 
The anterior half of the eyes was removed and four trials were performed on open-sky pig eyes. 
The eyes were mounted on a stage under a surgical OCT system, and a standard 48-gauge nano-cannula (MedOne Surgical) was mounted onto the robot and aligned with the sclera.\\
The procedure was divided into two phases. The first phase involved a far-field approach, where the cannula tip was advanced from outside the eye to a position approximately 1 mm above the retinal surface. This trajectory was planned and executed with smooth tool motion through the vitreous, avoiding retinal contact. In the second phase, the robot transitioned to a near-field mode using its fine motion control capabilities. In this phase, incremental motions on the order of 1 µm were used to approach the retinal surface and perform penetration to reach the RPE layer.\\
Throughout the procedure, OCT B-scans, volume scans, and camera imaging were used for tool-to-retina distance estimation and visual confirmation of subretinal bleb formation.


\subsection{Experimental Results}
Using the developed system, we performed subretinal injection attempts on six fresh ex vivo pig eyes. In each case, the robot executed the planned far-field and near-field motions in sequence, enabling stable tool insertion with micron-level precision. Of the six procedures, five resulted in successful subretinal bleb formation ($83\%$ success rate), confirmed via OCT imaging. The robot consistently maintained accurate trajectory control and demonstrated the ability to execute incremental near-field advancement with the required precision.
The one unsuccessful case did not result from any deficiency in the robotic control or precision. Instead, the cannula failed to penetrate the retinal surface, likely due to variability in tissue properties or dulling of the nanocannula over multiple trials. No erratic motion or trajectory deviation was observed during this trial.\\
Overall, these results validate the ability of the IRISS system to perform precise and repeatable subretinal injections. The combination of fine-resolution actuation, accurate RCM constraint, and stable near-field control positions the robot as a promising platform for advancing microsurgical techniques in ophthalmology.

\begin{figure}
    \centering
    \includegraphics[width=0.7\linewidth]{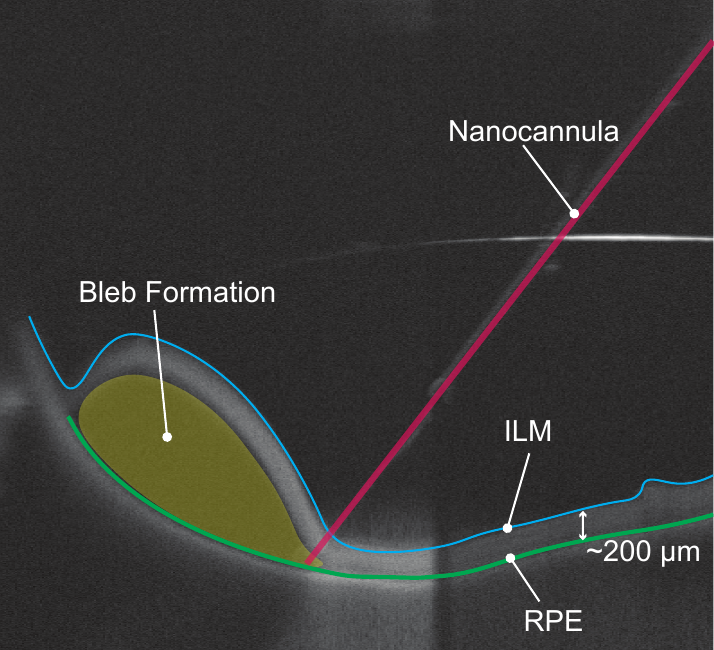}
    \caption{Shown is successful subretinal injection and bleb formation with the robotic system. The tool-tip is accurately positioned within the 10 \textmu m space between the ILM and RPE layers.}
    \label{fig:subretinal_injection}
\end{figure}

\section{Conclusion} \label{conclusion}

\noindent To reduce complications in intraocular procedures such as cornea incision stress, posterior capsule rupture, and retinal detachment, a precise robotic system was designed, manufactured, and evaluated.
Compared to other robotic systems, the developed robotic system features the following mechanical properties that are suitable for various types of intraocular procedures:
\begin{itemize}
    \item The ability to manipulate surgical instruments in a higher degrees of freedom.
    \item A magnetic lock-and-key mechanism enabling seamless tool exchange with minimum misalignment.
    \item Numerical optimization of the DH parameters that increases the reachable workspace.
\end{itemize}
The DH parameters were further calibrated with the use of OCT to reduce any misalignment during robot assembly.

The tooltip accuracy of the developed robotic system was evaluated by the OCT to be with a position error of 0.053$\pm$0.031 mm and an orientation error of 0.23$^\circ\pm$0.12$^\circ$.
In addition to the accuracy, the positional precision of the tooltip is 0.015$\pm$0.009 mm and the orientation precision of the tooltip is 0.048$^\circ\pm$0.027$^\circ$ across different surgical instruments.
The positional precision (0.133$\pm$0.062 mm) and orientation precision (0.161$^\circ\pm$0.075$^\circ$) of the tool exchange suggest that the magnetic lock-and-key mechanism is capable of accurately positioning the tool.
The incremental precision of 1 \textmu m further suggests that the robotic arm is further capable of performing delicate retina procedures such as retinal vein cannulation or sub-retinal injection.


\section*{Author Contribution}
Lai performed system integration, robot calibration, and experiments;
Rosen conceived robotic mechanism;
Foroutani performed subretinal injection experiments;
Ma performed mechanical design and motor control;
Wu performed robot calibration and experiments;
Hubschman contributed surgical insights and requirements;
Tsao led the project implementation;
The authors have equal contributions to the writing of the manuscript.

\bibliographystyle{IEEEtran}
\bibliography{IRISSv2}

\end{document}